\documentclass[letterpaper, 10 pt, journal, twoside]{ieeetran}
\usepackage{times}
\usepackage{multicol}
\usepackage{bm}
\usepackage[bookmarks=true]{hyperref}
\usepackage{graphicx}
\usepackage{amsmath}
\usepackage{amssymb}
\usepackage{caption}
\usepackage{pgfplots}
\usepackage{tikz}
\usepackage{pgfplots}
\usepackage{tikz-3dplot}
\usepackage{tikz-3dplot-circleofsphere}
\usepackage{circuitikz}
\usepackage{subfigure}
\usepackage{booktabs}
\usepackage{textcomp}
\usepackage{stfloats}
\usepackage{url}
\usepackage{verbatim}
\usepackage{bm}
\usepackage{cite}
\usepackage{color}
\usepackage{caption}
\usepackage{multirow}
\usepackage{booktabs}
\usepackage{amssymb}

\begin{document}

\markboth{IEEE Robotics and Automation Letters. Preprint Version. Accepted September, 2024}
{Li \MakeLowercase{\textit{et al.}}: Open-Structure}

\author{Yanyan Li$^{1}$, Zhao Guo$^{2}$, Ze Yang$^{2}$, Yanbiao Sun$^{2}$, Liang Zhao$^{3}$ and Federico Tombari$^{1,4}$
\thanks{$^{1}$Yanyan Li, Federico Tombari are with the Department of Computer Science, Technical University of Munich, Germany {\tt\footnotesize yanyan.li@tum.de, tombari@in.tum.de}}%
\thanks{$^{2}$Zhao Guo, Ze Yang, Yanbiao Sun are with the Tianjin University, China {\tt\footnotesize {(zeyang, guozhao, yanbiao.sun)}@tju.edu.cn}}%
\thanks{$^{3}$Liang Zhao is with Institute of Perception, Action, and Behaviour, School of Informatics, University of Edinburgh, UK {\tt\footnotesize liang.zhao@ed.ac.uk}}%
\thanks{$^{4}$Federico Tombari is with Google, Switzerland.}
}


\title{\LARGE \bf
Open-Structure: Structural Benchmark Dataset for SLAM Algorithms}

\maketitle

\begin{abstract}
\textcolor{black}{This paper presents Open-Structure, a novel benchmark dataset for evaluating visual odometry and SLAM methods. Compared to existing public datasets that primarily offer raw images, Open-Structure provides direct access to point and line measurements, correspondences, structural associations, and co-visibility factor graphs, which can be fed to various stages of SLAM pipelines to mitigate the impact of data preprocessing modules in ablation experiments.
The dataset comprises two distinct types of sequences from the perspective of scenarios. The first type maintains reasonable observation and occlusion relationships, as these critical elements are extracted from public image-based sequences using our dataset generator. In contrast, the second type consists of carefully designed simulation sequences that enhance dataset diversity by introducing a wide range of trajectories and observations.
Furthermore, a baseline is proposed using our dataset to evaluate widely used modules, including camera pose tracking, parametrization, and factor graph optimization, within SLAM systems.} By evaluating these state-of-the-art algorithms across different scenarios, we discern each module's strengths and weaknesses in the context of camera tracking and optimization processes.
The Open-Structure dataset and baseline system are openly accessible \textcolor{black}{on website: \url{https://open-structure.github.io}}, encouraging further research and development in the field of SLAM.
\end{abstract}


\section{INTRODUCTION}\label{section::Introduction}
\IEEEPARstart{C}{amera} pose estimation~\cite{mur2017orb,von2008lsd,qin2018vins} and scene reconstruction~\cite{wang2019real,di2019monocular,von2008lsd,di2020unified} 
play crucial roles as bridges for interaction between robots and unknown environments. These processes enable intelligent agents to determine their location and set the stage for advanced scene comprehension tasks. \textcolor{black}{While general SLAM systems~\cite{campos2021orb,qin2018vins,rosinol2020kimera} have recently achieved impressive performance levels}, they still have problems in challenging scenarios and fast camera motions. \textcolor{black}{To address these challenges}, algorithms focusing on specific modules, such as structural regularities~\cite{li2020structure,li2021rgb,li2022graph}, parameterization~\cite{civera2008inverse,zhao2011parallax}, and new optimization loss functions~\cite{rosinol2020kimera}, 
\textcolor{black}{have been proposed to pursue further improvements in this domain}.

Since these modules, \textcolor{black}{especially for SLAM back-end methods cannot directly utilize raw images for evaluation}, preprocessing steps to extract features, \textcolor{black}{match correspondences, and establish initial factor graphs are required first, which introduces uncertainties into the evaluation process, affecting the fairness and efficiency of the evaluations. Given the importance of fair and efficient evaluation in advancing the field, there is a growing need within the community for datasets that are carefully designed to meet the requirements of ablation evaluation for visual SLAM methods}.


\begin{figure}
    \centering
    \subfigure[\textcolor{black}{The ground truth scene that includes 3D points (red), individual lines (black) and groups of parallel lines (a group with the same color).}]{
        \resizebox{0.96\linewidth}{!}{\begin{tikzpicture}
                \node(img0) (0,0) {
                    \includegraphics[width=\linewidth]{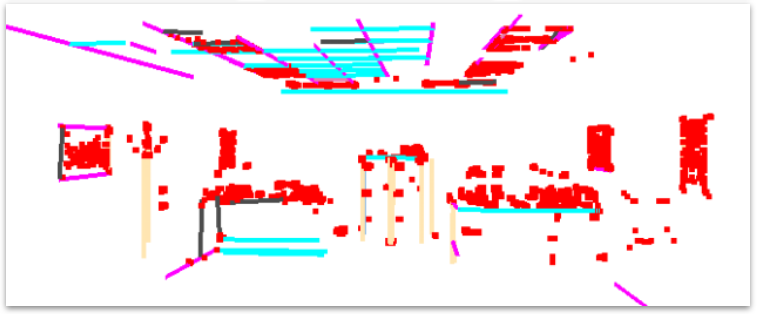}};
            \end{tikzpicture}}
            \label{subfig:teaser_a}}
    \subfigure[\textcolor{black}{\textcolor{black}{2D point and line measurements from three different viewpoints, with points in green, parallel line segments marked in the same color, and individual lines shown in black.}}]{
        \resizebox{.96\linewidth}{!}{\begin{tikzpicture}
                \node(img1) at (-.5,0) {
                    \includegraphics[width=.31\linewidth]{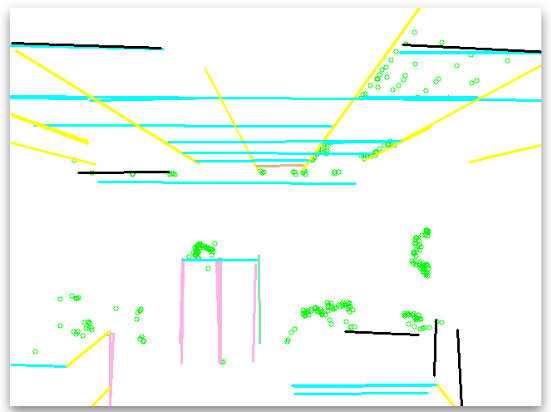}};
                \node(img2) at (2.2,0) {
                    \includegraphics[width=.31\linewidth]{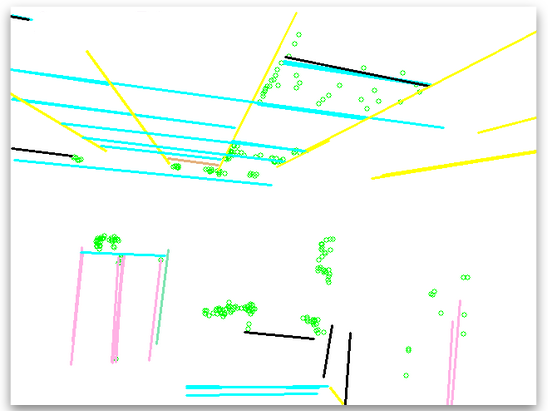}};
                \node(img3) at (4.9,0) {
                    \includegraphics[width=.31\linewidth]{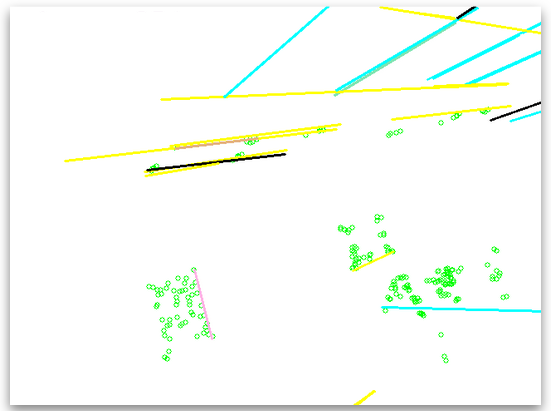}};
                \node [below,text width=1cm,align=center] at (img1.north){\tiny Frame 1};
                \node [below,text width=1cm,align=center] at (img2.north){\tiny Frame 51};
                \node [below,text width=1cm,align=center] at (img3.north){\tiny Frame 354};
            \end{tikzpicture}}
            \label{subfig:teaser_b}}
    \subfigure[Factor graph construction and optimization in the \textcolor{black}{proposed SLAM} baseline.]{
\includegraphics[width=0.49\linewidth, trim = 1.5cm 5.8cm 14.3cm 2.9cm, clip]{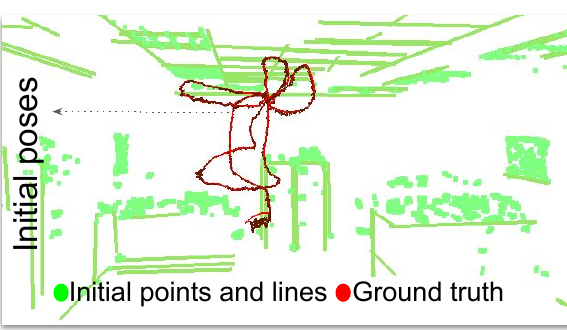}   
\includegraphics[width=0.49\linewidth, trim = 1.45cm .1cm 14.3cm 9.0cm, clip]{images/teaser_c12}  
            \label{subfig:teaser_c}
            }
    \caption{\textcolor{black}{Examples of data provided in Open-Structure Benchmark Dataset.}  }
    \label{fig:teaser}
\end{figure}

Public datasets~\cite{Sturm2012tum,handa2014ICL,tartanair2020iros} are \textcolor{black}{instrumental in advancing SLAM theories by reducing the reliance on high-precision hardware equipment and streamlining the evaluation process.}
Generally, we have real-world and synthetic datasets for visual SLAM tasks, where those real-world datasets, such as TUM RGB-D~\cite{Sturm2012tum}, EuRoC MAV~\cite{Burri2016EuRoC}, and KiTTi~\cite{Geiger2013KITTI}, are collected in real environments based on elaborate sensor setups. While synthetic ones, including ICL-NUIM~\cite{handa2014ICL}, Virtual-KiTTi~\cite{Gaidon2016Virtual}, and TartanAir~\cite{tartanair2020iros}, are generated from 3D rendering engines by designing virtual scenes and viewpoints. \textcolor{black}{Real-world datasets offer more realistic usage scenarios, but their collection process is expensive, and the properties of the hardware setup can impact the accuracy of the results. On the other hand, while synthetic sequences may lack realism compared to real-life data, they offer flexibility in providing various data formats of ground truth}, such as disparity, normals, and semantics.

\textcolor{black}{Datasets in raw-image style are well-suited for evaluating overall SLAM systems but are not directly applicable when evaluating SLAM back-end algorithms, such as parameterization approaches. This is because several preprocessing operations}, such as correspondence detection and association, are \textcolor{black}{typically} required to extract target information from raw images. Since ensuring that the preprocessed results generated from different SLAM pipelines are the same is complicated, \textcolor{black}{uncertainties will be incorporated into ablation studies. Therefore, simulation sequences are often designed to ensure consistent inputs in comparative experiments}.
\textcolor{black}{For example}, \cite{forster2016manifold} builds a cube environment with points randomly \textcolor{black}{distributed} on surfaces. The simulated visual observations, acceleration, and gyroscope measurements are used to evaluate on-manifold preintegration approaches for visual-inertial odometry.
~\cite{zhaooccupancy} designs two simulation scenes with obstacles and robot trajectories to evaluate occupancy map optimization. However, \textcolor{black}{these simulation datasets often oversimplify the generation process, lacking consideration for real observation relationships and occlusion issues, thus failing to simulate real-world scenarios accurately}.

\textcolor{black}{To enhance the efficiency of experimental comparisons in SLAM algorithms, this paper introduces a versatile benchmark dataset that provides point and line measurements, feature correspondences, structural relationships, and initial co-visibility factor graphs to the community. 
\textcolor{black}{The dataset comprises 22 sequences, divided into two parts, S-\uppercase\expandafter{\romannumeral1} and S-\uppercase\expandafter{\romannumeral2}. In the first part, reasonable observation and occlusion properties are preserved, as these $16$ sequences} are derived from raw images obtained from TUM RGB-D~\cite{Sturm2012tum}, ICL-NUIM~\cite{handa2014ICL} and TartanAir~\cite{tartanair2020iros} datasets. This preservation \textcolor{black}{introduced in Section~\ref{section::Measurements}} ensures that these sequences inherit more realistic spatial relationships.
    Additionally, \textcolor{black}{$6$ simulated sequences in the S-\uppercase\expandafter{\romannumeral2} group} enhance the dataset's diversity by introducing various carefully designed trajectories and scenes. Based on the proposed dataset, methods in areas such as initial pose estimation, parameterization, optimization, and loop closure adjustment can be evaluated directly, as depicted in Figure~\ref{fig:teaser}}.  
Furthermore, we provide a baseline architecture that is an incremental tracking and sparse mapping system with different initial pose estimation strategies, point and line representations, and optimization modules. By Feeding our point and line measurements, the baseline can estimate initial camera poses and build a sparse point-line map. Based on the given factor graph data, the baseline implements entrance for testing parameterization and optimization methods.
The contributions of this paper are summarized as
\begin{itemize}
    \item a benchmark dataset containing multiple data formats is provided for evaluating SLAM modules;
    \item a SLAM baseline is open-sourced to the community, providing tracking and optimization interfaces for using 2D \textcolor{black}{image} measurements and factor graph data;
    \item tracking, parameterization, and optimization modules are implemented and evaluated on our benchmark.
\end{itemize}

\section{Dataset Features}

\begin{table}
    \resizebox{\linewidth}{!}{
        \begin{tabular}{llc}
            \toprule
            \textbf{Scene type}                      & \textbf{Description}                            & \textbf{Equipped} \\ \hline
            S-\uppercase\expandafter{\romannumeral1} (16 sequences) & Reasonable observation and occlusion  & $\checkmark$                                                        \\
            S-\uppercase\expandafter{\romannumeral2} (6 sequences) & Simulation trajectories and scenes & $\checkmark$      \\ \hline \hline
            \textbf{Data type}                       & \textbf{Denotation}                             & \textbf{Equipped} \\ \hline
            Point measurements                       & \begin{tabular}[l]{@{}l@{}}\textcolor{black}{$\mathbf{p}=[\mathbf{u}^T\; d]^T$}\\ pixel position and depth\end{tabular}
                                                     & $\checkmark$                                                        \\ \hline
            Line measurements                        & \begin{tabular}[l]{@{}l@{}}$\mathbf{l}=\left[{\mathbf{p}^s}^T \; {\mathbf{p}^e}^T\right]^T$\\ 2D start point and end point\end{tabular}                   & $\checkmark$      \\ \hline
            Point landmarks                          & \begin{tabular}[l]{@{}l@{}}$\mathbf{P}_w=\left[ X_w\; Y_w\; Z_w \right]^T$ \\  3D point in the world coordinates \end{tabular}                   & $\checkmark$      \\ \hline
            Line landmarks                           & \begin{tabular}[l]{@{}l@{}}$\mathbf{L}_w=\left[{\mathbf{P}^s_w}^T \; {\mathbf{P}^e_w}^T\right]^T$ \\  3D endpoints in the world coordinates \\
                $\mathbf{\mathcal{L}}_w = \left[
                            {\mathbf{n}_w}^T \; {\mathbf{d}_w}^T \right]^T$ \\
                \textit{Pl\"ucker Representation}                                     \\
                $\bm{\mathcal{O}}_w = [\bm{\phi}^T_w \; \varphi_w]^T$                 \\
                \textit{Orthonormal Representation}                                   \\
            \end{tabular}                  & $\checkmark$      \\ \hline
            Correspondences                          &
            \begin{tabular}[l]{@{}l@{}}$\mathfrak{C}=[\mathbf{p}^i_{c_0},\dots,\mathbf{p}^i_{c_j}]$\\ \textcolor{black}{$\mathbf{p}^i_{c_0},\dots,\mathbf{p}^i_{c_j}$ are associated with $\mathbf{P}^i_w$} \end{tabular}           & $\checkmark$                                                        \\ \hline
            Structural 3D Lines                      & \begin{tabular}[l]{@{}l@{}}$\mathfrak{S}=[\mathbf{L}^0_w,\dots,\mathbf{L}^k_w]$ \\ \textcolor{black}{$\mathfrak{S}$ is a group of 3D parallel lines}
            \end{tabular}
                                                     & $\checkmark$                                                        \\
            \bottomrule
        \end{tabular}}
    \caption{\textcolor{black}{Overview of the Open-Structure benchmark dataset. There are two subsets of sequences, S-\uppercase\expandafter{\romannumeral1} and S-\uppercase\expandafter{\romannumeral2}, where sequences in S-\uppercase\expandafter{\romannumeral1} and S-\uppercase\expandafter{\romannumeral2} are generated based on image-based public datasets and self-designed simulation scenes, respectively. Additionally, the types of data provided by each sequence include measurements, landmarks, correspondences, and structural relationships. 
    }}
    \label{tab:features_benchmark}
\end{table}

As listed in Table~\ref{tab:features_benchmark}, Open-Structure provides multi-model data, such as measurements (points and lines), initial landmarks and poses, associated structural lines, and co-visibility factor graphs. \textcolor{black}{The dataset also includes ground truth poses and 3D landmarks, enabling the evaluation of estimated trajectories and reconstruction results}.


\begin{figure}
    \centering
    \subfigure[\textcolor{black}{Statistics of the number of point and line features on each image.}]{
            \begin{tikzpicture}
                \begin{axis}[
                        ylabel=\textcolor{black}{Number\;of\;features},
                        xlabel={Frame},
                        label style={font=\small},
                        width=0.92*4.5*1.5cm,
                        height=0.92*2.5*1.5cm,
                        scale only axis,
                        enlargelimits=false,
                        axis on top,
                        xmin=0, xmax=100,
                        scaled x ticks=false,
                        xtick={0,10,...,100},
                        ymin=0, ymax=140,
                        ytick={0,20,...,140}]
                    \addplot graphics[xmin=0,xmax=100,ymin=0,ymax=140] {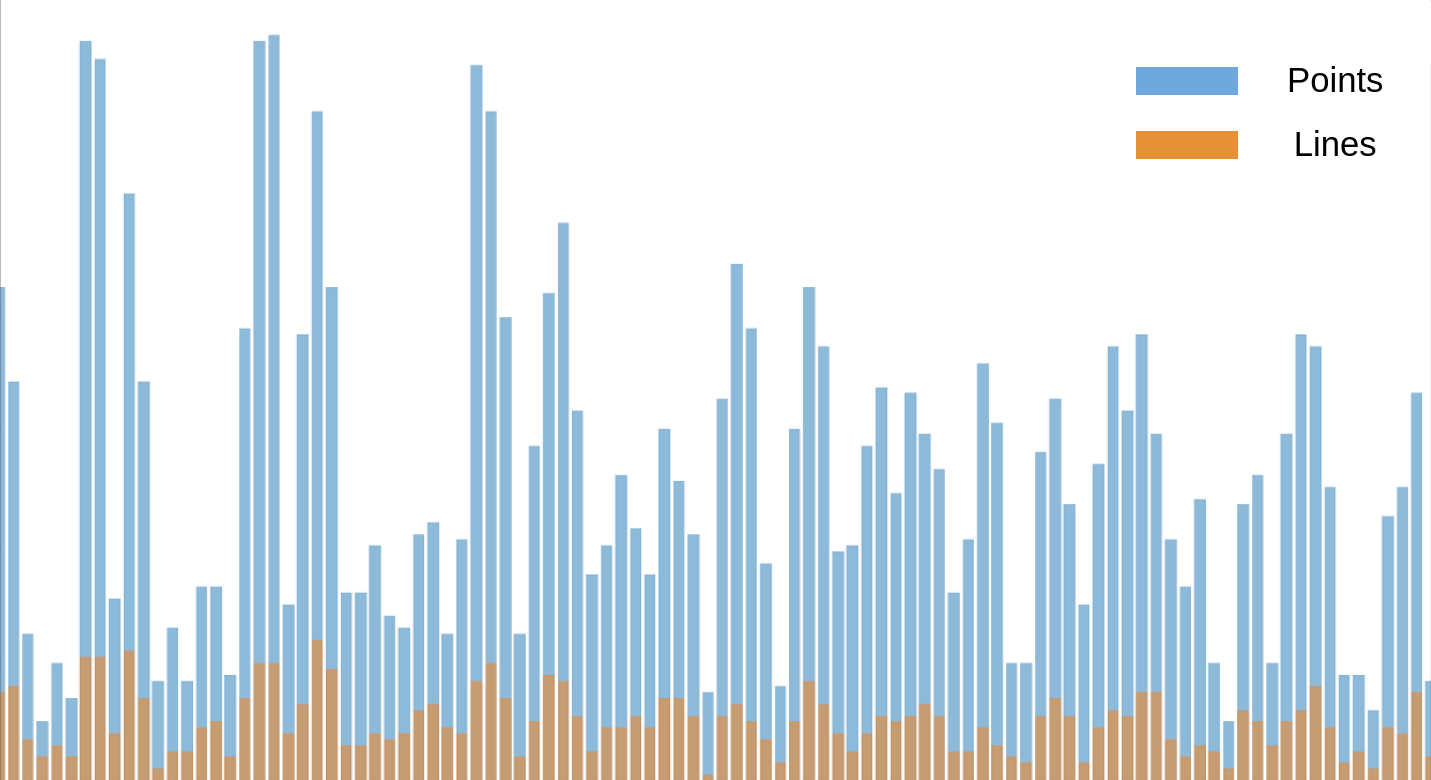};
                \end{axis}
            \end{tikzpicture}
            \label{subfig:meas}
        }
    \subfigure[\textcolor{black}{Statistics of the number of occupied pixel cells on each image. An occupied cell means that one or more features are located within a cell.}]{
            \begin{tikzpicture}
                \begin{axis}[
                        ylabel=\textcolor{black}{Number\;of\; occupied\;cells},
                        xlabel={Frame},
                        label style={font=\small},
                        width=0.92*4.5*1.5cm,
                        height=0.92*2.5*1.5cm,
                        scale only axis,
                        enlargelimits=false,
                        axis on top,
                        xmin=0, xmax=100,
                        scaled x ticks=false,
                        xtick={0,10,...,100},
                        ymin=0, ymax=160,
                        ytick={0,20,...,160}]
                    \addplot graphics[xmin=0,xmax=100,ymin=0,ymax=160] {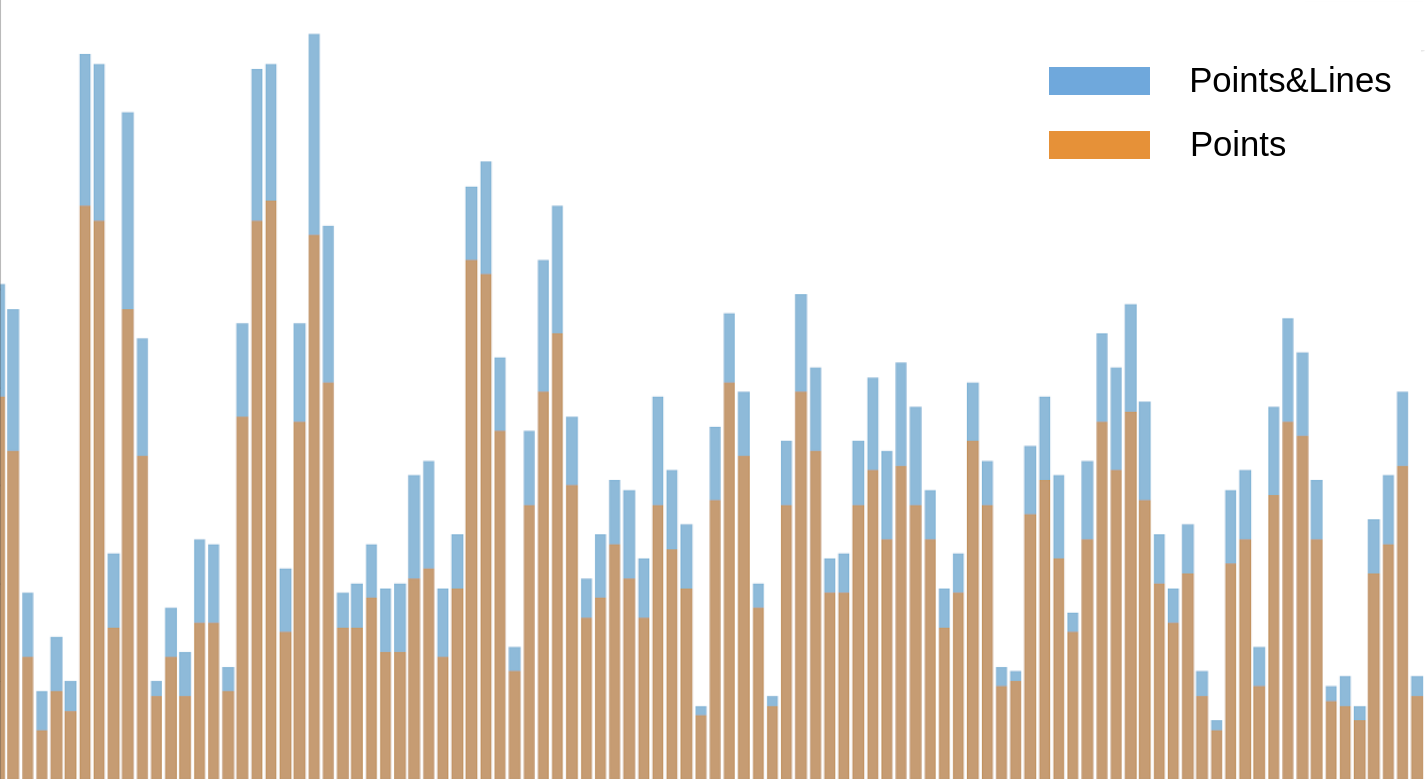};
                \end{axis}
            \end{tikzpicture}
            \label{subfig:distribution}
        }
    \caption{\textcolor{black}{Statistical analysis per frame of the \textit{sphere2} sequence.}}
    \label{fig:feature_size}
\end{figure}

\subsection{Overview of Scene and Data Types}

\textcolor{black}{The proposed dataset consists of two parts of sequences. The first part (S-\uppercase\expandafter{\romannumeral1}) comprises 16 sequences established based on image-based publicly available datasets~\cite{handa2014ICL,Sturm2012tum,tartanair2020iros}. In contrast, the second part (S-\uppercase\expandafter{\romannumeral2}) consists of 6 sequences generated by the simulation module of our baseline. Despite the different sources used to create these two parts, the data format of these sequences is standardized}.


\textcolor{black}{For each sequence, ground truth poses and 3D landmarks are provided, as depicted in Figure~\ref{fig:teaser}(a). The dataset includes 3D parallel lines, implying additional structural relationships among the 2D line segments, as shown in Figure~\ref{fig:teaser}(b). Using correspondences between frames, the baseline estimates relative poses through frame-to-frame pose computation}. \textcolor{black}{By utilizing initial camera poses and measurements}, the baseline maintains a local map to manage estimated point and line landmarks. This map-to-frame \textcolor{black}{tracking} strategy is also implemented \textcolor{black}{to estimate} camera poses. After obtaining observation relationships and estimating initials, the co-visibility factor graphs are constructed in the baseline, \textcolor{black}{as illustrated in Figure~\ref{subfig:teaser_c}}, which can be optimized based on points and lines. 
In the proposed benchmark dataset, all those mentioned data types, \textbf{point and line measurements}, \textbf{structural 2D and 3D lines}, \textbf{3D-2D observations}, and \textbf{co-visibility factor graphs}, can be directly obtained. Section~\ref{sec:meas_generate} and \ref{sec:initial_estimate}, respectively, introduce the details of \textcolor{black}{measurement} generation and initial estimation process.

\subsection{Statistical Analysis}
Traditionally, \textcolor{black}{visual SLAM sequences are described using qualitative characteristics, such as low-textured or high-textured, structured or unstructured, as shown in the TUM RGB-D dataset~\cite{Sturm2012tum}}. However, in our dataset, each sequence's profile can be characterized by quantitative values. These include the number of features in each frame, feature distributions based on occupied grids, and the number of structural constraints. For example, the \textit{sphere2} sequence contains 100 images, as depicted in Figure~\ref{fig:feature_size} which also illustrates the number of 2D features detected in each image. These quantitative statistics allow for a more precise understanding of each sequence.



\textcolor{black}{To analyze feature distributions, we divide each image into cells with a size of $10\times{10}$ pixels. If any part of a feature falls within a cell, we consider the cell occupied. This method applies to both points and line endpoints. Figure~\ref{subfig:distribution} illustrates the number of occupied cells in each image, providing a visual representation of the sequence's feature distribution.
By combining the statistical results of feature numbers and feature distributions from Figure~\ref{fig:feature_size}, we observe that while many points tend to cluster together in a few cells, cells occupied by endpoints of lines help improve distribution uniformity}.



\section{Measurement Generation}\label{sec:meas_generate}
This section introduces the process of \textcolor{black}{generating measurements} of our dataset, as described in Table~\ref{tab:features_benchmark}, the measurement of each point \textcolor{black}{$\mathbf{p}$ (or $\mathbf{l} = [{\mathbf{p}^s}^T\; {\mathbf{p}^e}^T]^T$ for line feature endpoints)} contains the pixel position \textcolor{black}{$\mathbf{u}=[p_x\;p_y]^T$} and depth value $d$, which \textcolor{black}{are applicable in} monocular, stereo, and RGB-D SLAM methods. 

\textcolor{black}{For sequences in S-\uppercase\expandafter{\romannumeral1}, 3D scenes} (see Section~\ref{section::Measurements}) are generated based on RGB-D sequences in ICL-NUIM~\cite{handa2014ICL}, TUM RGB-D~\cite{Sturm2012tum}, and TartanAir~\cite{tartanair2020iros} \textcolor{black}{datasets}. Starting from RGB-D images and ground truth poses, \textcolor{black}{these sequences provide more realistic spatial characteristics}, such as scene layouts, occlusion, and robot trajectories. To increase the diversity of trajectories and observation types, simulation sequences (see Section~\ref{sec:meas_simulate}) are generated in designed trajectories and environments, where sharp/medium/slow rotational movements and rich/medium/low observations are considered to evaluate the performance of SLAM methods in extreme cases.


\begin{figure}
    \centering
    \includegraphics[width=0.95\linewidth]{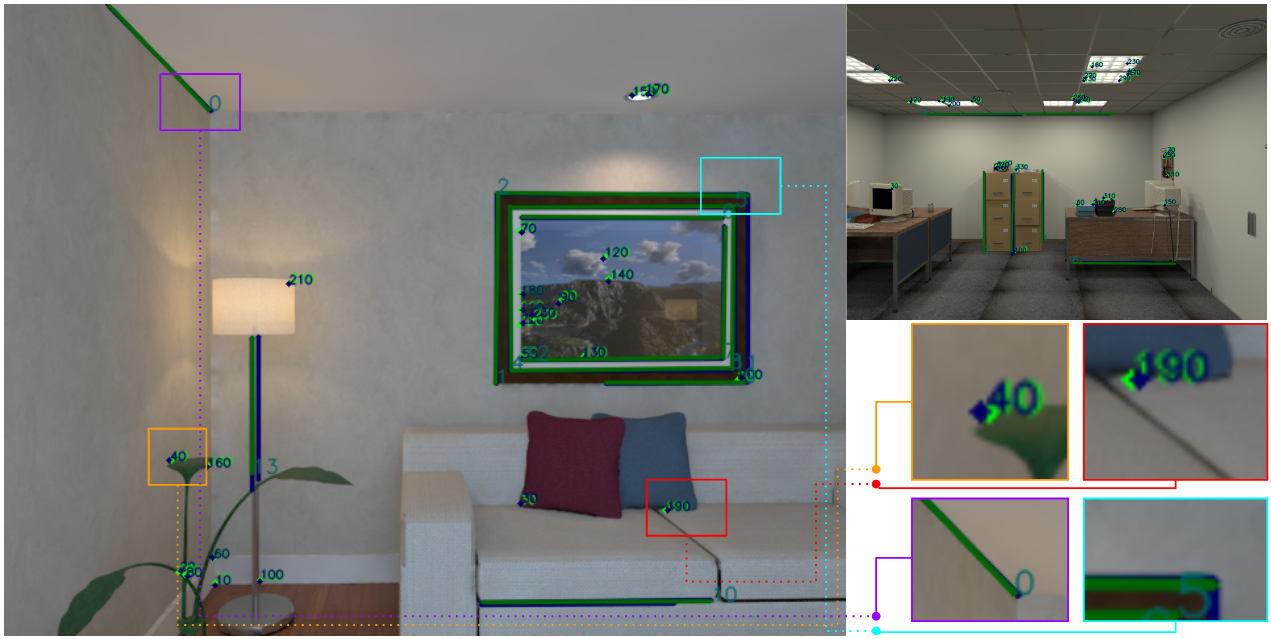}
    \caption{Comparison in positions of detected features (blue) and \textcolor{black}{generated} measurements (green) of our dataset.}
    \label{fig:orb_ours}
\end{figure}

\subsection{Observation Association based on RGB-D Inputs}\label{section::Measurements}
\textcolor{black}{Given a pair of RGB and depth images}, ORB~\cite{Rublee2011ORB} and FLD~\cite{lee2014outdoor} detectors are used to extract 2D points and line segments. \textcolor{black}{For every point feature}, pixel position \textcolor{black}{$\mathbf{u}$} and depth $d$ reconstruct a point landmark $\mathbf{P}_{c_j}$ in the $j^{th}$ camera coordinate frame based on the intrinsic matrix $\mathbf{K}$. And $\mathbf{P}_{c_j}$ is transformed to $\mathbf{P}_{w}$, in the world coordinates, via ground truth camera pose $\mathbf{T}_{w,c_j}$. Because of noise in this process, \textcolor{black}{the 3D back-projections of those correspondences may not align perfectly in 3D space}. Therefore, \textcolor{black}{these 3D points are fused into one landmark first}, and then the observation association step detects all observation relationships between the landmark and its observations in images. Based on the KD-Tree approach, when a new map point is generated from the current image, it is merged into the map by considering distance and descriptor thresholds. New landmarks are added to the map directly, while existing ones are associated with corresponding landmarks. After association, the $\mathbf{P}_{w}$ will be re-projected to the image to obtain the ground truth observations.

\begin{figure*}
    \centering
    \subfigure[\textit{office0}]{
        \includegraphics[width=0.48\textwidth]{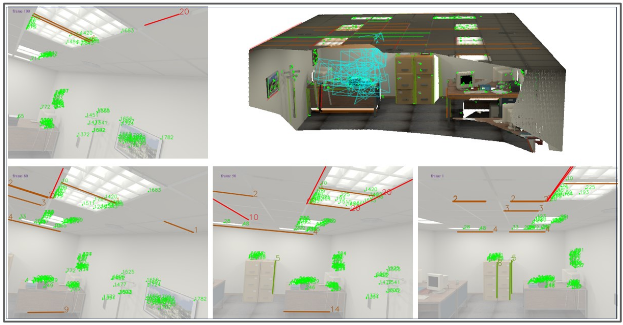}}
    \subfigure[\textit{box1}]{
        \label{fig:simu_scene}
        \includegraphics[width=0.48\textwidth]{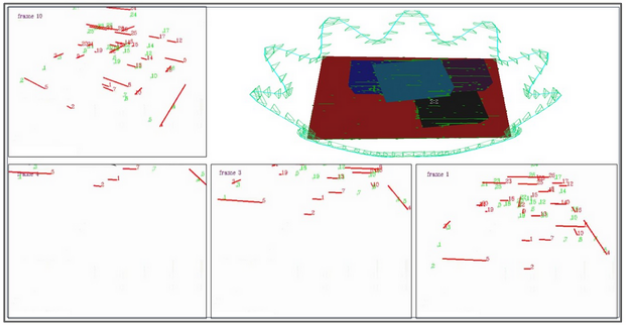}}
    \caption{Overview of trajectories, 3D models and 2D measurements of Open-Structure sequences. Point and line measurements are colored in green and red, respectively. \textcolor{black}{(a) \textit{office0} comes from S-\uppercase\expandafter{\romannumeral1} showing reasonable feature distributions. (b) \textit{box1} comes from S-\uppercase\expandafter{\romannumeral2} presenting more challenging trajectories and observations}.}
    \label{fig:data_img_overview}
\end{figure*}



For each 2D line segment extracted from the image, discrete pixels lying on the line are \textcolor{black}{back-projected} to the world coordinate frame, similar to the process for point landmarks, using depth values, $\mathbf{K}$, and $\mathbf{T}_{w,c_j}$. These discrete 3D points are then used in the RANSAC model to fit an endpoint-based line representation, \textcolor{black}{$\mathbf{L}_{w} = \left[ \begin{array}{cc}
            {\mathbf{P}_{w}^{s}}^T {\mathbf{P}_{w}^{e}}^T 
        \end{array}\right]^T$}, in the world coordinate frame. Observation relationships between 2D line features and the landmark $\mathbf{L}_{w}$ \textcolor{black}{are established}.         
The map line fusion approach involves comparing the angles of direction vectors and line distances between existing map lines and newly reconstructed ones to establish the co-visible line relationship between different frames. During the map line fusion process, the positions of line landmarks are updated incrementally. Since we only record the observed relationships in the incremental process, the observations in pixel and depth values are computed when we finish the updating process. 
%
\textcolor{black}{To establish structural relationships within our dataset, we classify 3D lines into distinct groups of parallel lines based on the angle threshold between their direction vectors. As the selected 3D lines within each group are not perfectly parallel, we first unify their directions and then calculate the average direction vector to build the new direction for the group. By reprojecting the endpoints onto this newly obtained direction, all these lines become perfectly parallel}.


While the positions of 3D point and line landmarks are updated in the fusion process, these changes are generally small. \textcolor{black}{As shown in Figure~\ref{fig:orb_ours}, the re-projections of updated landmarks remain very close to the original positions computed by feature detection methods, which  demonstrates that the data provided by our dataset exhibits reasonable feature distributions. It is important to note that the observations described here are not actual measurements}. The noise model that converts observations to measurements is introduced in Section~\ref{sec:noise_model}.


\subsection{Observation Association based on Simulator}~\label{sec:meas_simulate}
In the simulated sequences, \textcolor{black}{RGB-D images are not used in measurement generation}. \textcolor{black}{Instead}, 3D scenes and trajectories are designed first, and sparse landmarks are randomly distributed on the surface of objects. \textcolor{black}{The 3D scene is constructed using several cube boxes, as shown in Figure~\ref{fig:simu_scene}, and each view in the wave-shape trajectory focuses on the scene. Observations are obtained by re-projecting these landmarks to each frame, with points and lines being considered if they lie on the image planes of those views. Similarly, in this process, the re-projected pixels are converted to measurements by adding 2D Gaussian noise}.

\begin{figure*}
    \centering
    \includegraphics[width=\linewidth, trim = 2 175 50 40, clip]{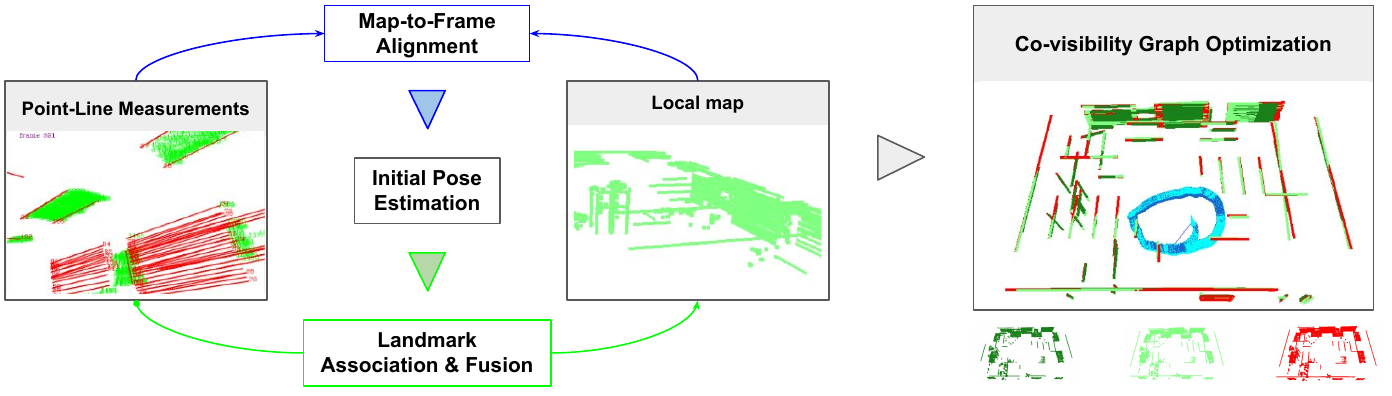}
    \caption{The architecture of the Open-Structure baseline that reads point and line measurements directly. Initial camera poses, and a sparse map are estimated simultaneously via 3D-2D alignment and landmark fusion blocks. Co-visibility observations, initial poses, and landmarks are fed to the Co-visibility Graph Optimization module, where \textcolor{black}{initial, optimized}, and ground truth landmarks are highlighted in dark green, light green, and red, respectively.}
    \label{fig:baseline_architecture}
\end{figure*}

\subsection{Noise Model for Measurements}\label{sec:noise_model}
\textcolor{black}{We introduce a noise model to transform \textcolor{black}{ground truth observations} into measurements in this section. For a 3D map point $\mathrm{P}_w$, the ground truth observation in the $j^{th}$ image is represented as \textcolor{black}{$[\hat{\mathbf{u}}, \hat{d}]$}. The noise model~\cite{handa2014ICL} is used to add noise to pixel positions and depth values to simulate the measurements}. 

\textcolor{black}{For the pixel position measurement \textcolor{black}{$\mathbf{u}$}, noise is added as \textcolor{black}{$\mathbf{u} = \mathbf{\hat{u}} + \bm{\alpha}$}, \textcolor{black}{where $\bm{\alpha} = [\alpha_x\; \alpha_y]^T$} is generated from 2D zero-mean Gaussian noise $\mathcal{N}(\mathbf{0}, \sigma^2_s\cdot \mathbf{I})$}. For depth measurements, we follow the process used in ICL-NUIM~\cite{handa2014ICL}. First, we build a disparity map based on the baseline model ($m=35130$) of the Kinect sensors~\cite{barron2013intrinsic}, and then add 2D Gaussian noise to the disparity map. \textcolor{black}{The noisy disparity map is then used to build the depth measurements. This process is expressed as
\begin{equation}
    d = \frac{m}{m/(\hat{d}+\alpha_d) + 0.5}
\end{equation}
here $\alpha_d \in \mathcal{N}(0, \sigma^2_d)$ simulates the noise generated by the noisy disparity map. $\sigma_s$ and $\sigma_d$ are set to $1$ and $1/6$, respectively}. The endpoints of lines are passed to the same process to obtain measurements.

\section{Baseline for initialization and optimization}\label{sec:initial_estimate}
\subsection{Initial Landmarks and Camera Poses}
\textcolor{black}{The initial pose estimation and sparse mapping are implemented incrementally via the proposed baseline architecture, as illustrated in Figure~\ref{fig:baseline_architecture}}. First, when a new set of measurements is inputted into the system, a sparse 3D map based on point and line landmarks is initialized. In the subsequent tracking process, the map is used to estimate the current camera pose $\mathbf{T}_{c_j,w}$ using the map-to-frame alignment module, \textcolor{black}{which is based on the EPnP method~\cite{lepetit2009ep}}. After obtaining the initial camera pose, the measurements from the current frame are fused into the sparse model. Co-visible measurements are fused to update the map, while new detections are initially reconstructed and merged into the map.

\begin{figure}
    \centering
    \resizebox{\linewidth}{!}{
        \begin{circuitikz}
            \tikzstyle{every node}=[font=\LARGE]
            \draw [short] (16.5,18.25) -- (16.5,16.5);
            \draw [short] (16.5,18.25) -- (17.5,17.5);
            \draw [short] (16.5,16.5) -- (17.5,15.75);
            \draw [short] (17.5,17.5) -- (17.5,15.75);
            \draw (14,17.25) to[short, -*] (14,17.25);
            \draw [ color={rgb,255:red,211; green,215; blue,207}, dashed] (14,17.25) -- (16.5,18.25);
            \draw [ color={rgb,255:red,211; green,215; blue,207}, dashed] (14,17.25) -- (16.5,16.5);
            \draw [ color={rgb,255:red,211; green,215; blue,207}, dashed] (14,17.25) -- (17.5,15.75);
            \draw [ color={rgb,255:red,211; green,215; blue,207}, dashed] (14,17.25) -- (17.5,17.5);
            \draw [->,color=blue ] (14,17.25) -- (14,18.5);
            \draw [->,color={red}] (14,17.25) -- (15.5,17.25);
            \draw [->,color= green] (14,17.25) -- (15,18.25);
            \draw [color={rgb,255:red,245; green,121; blue,0}](16.75,16.5) to[short, -*] (16.75,16.5);
            \draw [ color={rgb,255:red,193; green,125; blue,17}, short] (16.75,16.5) -- (17.25,16.1);
            \draw [ color={rgb,255:red,193; green,125; blue,17}, short] (16.75,16.5) -- (17.25,17.25);
            \draw [color={rgb,255:red,92; green,53; blue,102}](16.75,17.75) to[short, -*] (16.75,17.75);
            \draw [dashed] (14,17.25) -- (19.5,18.25);
            \draw [dashed] (14,17.25) -- (20.25,15.5);
            \draw [dashed] (14,17.25) -- (20.25,17.25);
            \draw [dashed] (14,17.25) -- (19.5,15.25);
            \draw [color={rgb,255:red,143; green,89; blue,2}](20.25,17.25) to[short, -*] (20.25,17.25);
            \draw [ color={rgb,255:red,143; green,89; blue,2}, ](20.25,17.25) to[short] (20.25,15.5);
            \draw (20.25,15.5) to[short, -*] (20.25,15.5);
            \draw [color={rgb,255:red,143; green,89; blue,2}](20.25,15.5) to[short, -*] (20.25,15.5);
            \draw [color={rgb,255:red,143; green,89; blue,2}](19.5,15.75) to[short, -*] (19.5,15.75);
            \draw [ color={rgb,255:red,143; green,89; blue,2}, ](19.5,15.75) to[short] (19.5,15.25);
            \draw [color={rgb,255:red,143; green,89; blue,2}](19.5,15.25) to[short, -*] (19.5,15.25);
            \draw [color={rgb,255:red,92; green,53; blue,102}](19.5,18.25) to[short, -*] (19.5,18.25);
            \node [font=\LARGE] at (20.9,16.75) {$\mathbf{\mathcal{L}}^k_w$};
            \node [font=\LARGE] at (20,15.0) {$\mathbf{\mathcal{L}}^{k+1}_w$};
            \node [font=\LARGE] at (20,18.25) {$\mathbf{P}^i_w$};
            \node [font=\normalsize] at (17.2,16.75) {$\mathbf{l}^k_{c_j}$};
            \node [font=\normalsize] at (16.6,16.00) {$\mathbf{l}^{k+1}_{c_j}$};
            \node [font=\normalsize] at (17.2,18.2) {\textcolor{black}{$\mathbf{u}^i_{c_j}$}};
            \draw (10.5,15.25) to[short, -*] (10.5,15.25);
            \draw [->,color=blue] (10.5,15.25) -- (10.5,16.5);
            \draw [->,color= red ] (10.5,15.25) -- (9.5,14.75);
            \draw [->,color= green] (10.5,15.25) -- (11.75,15.25);
            \node [font=\LARGE] at (10.3,16.7) {$z$};
            \node [font=\LARGE] at (9.5,15.1) {$x$};
            \node [font=\LARGE] at (11.75,15.5) {$y$};
            \node [font=\LARGE] at (11,15.75) {$W$};
            \node [font=\LARGE] at (13.9,16.8) {$c_{j}$};
            \draw [ color={rgb,255:red,239; green,41; blue,41}, ->] (11.5,16) .. controls (12,16.75) and (12.25,16.75) .. (13,16.75);
            \node [font=\LARGE] at (11.5,17.25) {$\mathbf{T}_{c_j,w}$};
        \end{circuitikz}}
    \caption{\textcolor{black}{$\mathbf{T}_{c_j,w}$ is the pose of the world coordinates $W$ with respect to the camera frame. Landmarks $\mathbf{\mathcal{L}}^k_w$, $\mathbf{\mathcal{L}}^{k+1}_w$ and $\mathbf{P}^i_w$ are detected by the $j^{th}$ camera, corresponding measurements on the image plane are $\mathbf{l}^k_{c_j}$, $\mathbf{l}^{k+1}_{c_j}$, and \textcolor{black}{$\mathbf{u}^i_{c_j}$}.}}
    \label{fig:reproject}
\end{figure}

\subsection{Co-visibility Factor Graph Construction}

Based on the estimated initial camera poses and landmarks, and measurements from the sequences, co-visibility factor graphs are constructed for further optimization. The vertices in this graph $\mathcal{G}$ contain camera poses, point landmarks, and line landmarks. To be specific, camera pose $\bm{T}_{c_j,w} = \left[\begin{array}{cc}
            \mathbf{R}_{c_j,w} & \mathbf{t}_{c_j,w} \\
            \mathbf{0}         & 1
        \end{array}\right]$, where $\mathbf{T}_{c_j,w} \in SE(3)$, $\mathbf{R}_{c_j,w}\in SO(3)$ and $\mathbf{t}_{c_j,w} \in \mathbb{R}^3$. Points used in the optimization module are represented as $\mathbf{P}^i_w = \left[\begin{array}{ccc}
            X^i_w & Y^i_w & Z^i_w \\
        \end{array}\right]^T$, and line landmarks are represented in Pl\"ucker Parametrization~\cite{bartoli2005structure} as $\bm{\mathcal{L}}^k_w = \left[\begin{array}{c}
            \bm{n}^k_w \\ 
            \bm{d}^k_w
        \end{array}\right]$ where $\mathbf{n}^k_w/||\mathbf{n}^k_w||$ is the unit normal vector of the plane built by the global origin and endpoints of the $k^{th}$ line, and $\mathbf{d}_w^k/||\mathbf{d}_w^k||$ is the unit direction vector of the 3D line. Since the Pl\"ucker representation has over-parameterization issues, \textcolor{black}{the orthonormal method~\cite{bartoli2005structure} is used to represent the lines in the iterative optimization steps}.

\section{Experiments}
In this section, initial camera pose estimation based on Frame-to-Frame (\textit{FtF}) and Map-to-Frame (\textit{MtF}) strategies, Point-BA~\cite{mur2017orb} and PointLine-BA~\cite{he2018pl} optimization methods are evaluated on the proposed dataset.

\subsection{Approaches in Experiments}
Optimization approaches evaluated in this paper include different optimization graph architectures and parametrization approaches. In the architecture of Co-visibility Factor Graph (CFG), we evaluate optimization strategies used in Point-BA~\cite{mur2017orb} and PointLine-BA~\cite{he2018pl} as listed in Table~\ref{tab:quantitative_results}.


\paragraph{Point-BA} following ORB-SLAM2~\cite{mur2017orb}, Point-BA represents point landmarks in Euclidean \textit{XYZ} form and uses a loss function based on re-projection errors between re-projected points and measurements.
Based on the point feature measurement model, \textcolor{black}{the measurement of the $i^{th}$ global point landmark $\mathbf{P}^i_w $ at frame $c_j$ is denoted as \textcolor{black}{$\mathbf{u}_{c_j}^{i} = [\begin{array}{cc}
        p_{x,c_j}^{i} & p_{y,c_j}^{i}
    \end{array}]^T$} as shown in Figure~\ref{fig:reproject}}, and the re-projection factor of a point feature is defined as \textcolor{black}{$\bm{r}_{\bm{p}}(\mathbf{u}_{c_j}^{i}, \mathbf{P}^i_w, \mathbf{T}_{c_j,w})$}, \textcolor{black}{where
\begin{equation}
        \bm{r}_{\bm{p}}(\mathbf{u}_{c_j}^{i}, \mathbf{P}^i_w,\mathbf{T}_{c_j,w})
        = \mathbf{u}_{c_j}^{i} - \Pi(\mathbf{P}^i_w,\mathbf{T}_{c_j,w})
    \label{r_p}
\end{equation} 
here $\Pi(\cdot)$ is the re-projection function}.

The target function is obtained by feeding the residual function (Equation~\ref{r_p}) into the non-linear least squares model~\cite{li2020co}. The corresponding Jacobian matrices $\bm{J}_{\bm{P}}$ and $\bm{J}_{\mathcal{X}}$ of the target function are obtained \textcolor{black}{to update} landmarks and camera states, respectively.

\paragraph{PointLine-BA} following PL-SLAM~\cite{he2018pl}, the Pl\"ucker presentation is used for re-projection error computation, and the Orthonormal method is employed for iterative optimization steps.
\input{figure.tex}
\textcolor{black}{The endpoint measurements of the $k^{th}$ global line landmark $\bm{\mathcal{L}}_w^k$ at frame $c_j$ are \textcolor{black}{$\mathbf{u}^{k,s}_{c_j}$} and \textcolor{black}{$\mathbf{u}^{k,e}_{c_j}$}. The distances between the re-projected line $\mathbf{l}^k_{c_j}$ and these two endpoint measurements are given by
\begin{equation}
\bm{r}_{\bm{l}}( \mathbf{u}^{k,s}_{c_j},\mathbf{u}^{k,e}_{c_j}, \bm{\mathcal{L}}_w^k, \mathbf{T}_{c_j,w} ) = \left[\begin{array}{c}
            d( \mathbf{u}^{k,s}_{c_j}, \mathbf{l}^k_{c_j} ) \\
            d( \mathbf{u}^{k,e}_{c_j}, \mathbf{l}^k_{c_j} )
        \end{array}\right]
\end{equation}
where \textcolor{black}{$d(\mathbf{u}^{k,s}_{c_j}, \mathbf{l}^k_{c_j})= \frac{{\mathbf{u}^{k,s}_{c_j}}^{T}\cdot \mathbf{l}^k_{c_j}}{\sqrt{l_0^2+l_1^2}}$}, and $\mathbf{l}^k_{c_j} = \left[\begin{array}{ccc}
            l_0 & l_1 & l_2\end{array}\right]^T$ is the 2D line re-projected from the $k^{th}$ 3D map line,} where the 2D line can be obtained by the cross product of homogeneous coordinates \textcolor{black}{$\mathbf{\bar{u}}^{k,s}_{c_j}$} and \textcolor{black}{$\mathbf{\bar{u}}^{k,e}_{c_j}$} of two re-projected endpoints via the following formulation: 
\begin{equation}
   \mathbf{l}^k_{c_j} = \frac{\mathbf{\bar{u}}^{k,s}_{c_j}\times \mathbf{\bar{u}}^{k,e}_{c_j}}{||\mathbf{\bar{u}}^{k,s}_{c_j}\times \mathbf{\bar{u}}^{k,e}_{c_j}||}
\end{equation} 
here $||\cdot||$ is the normalization opration and the re-projected endpoint can be obtained \textcolor{black}{via the $\Pi(\cdot)$ function}.
%

\subsection{Metrics}

All experiments are conducted on an Intel Core i7-8700 CPU (@3.20GHz). We evaluate the camera pose computation approaches on various sequences using Absolute Trajectory Errors (ATE) and Relative Pose Errors (RPE) to measure the absolute and relative pose differences between estimated and ground truth motions.


\begin{figure}
    \centering
    \includegraphics[width=\linewidth, trim = 3.7cm 3.2cm 5.6cm 3.2cm, clip]{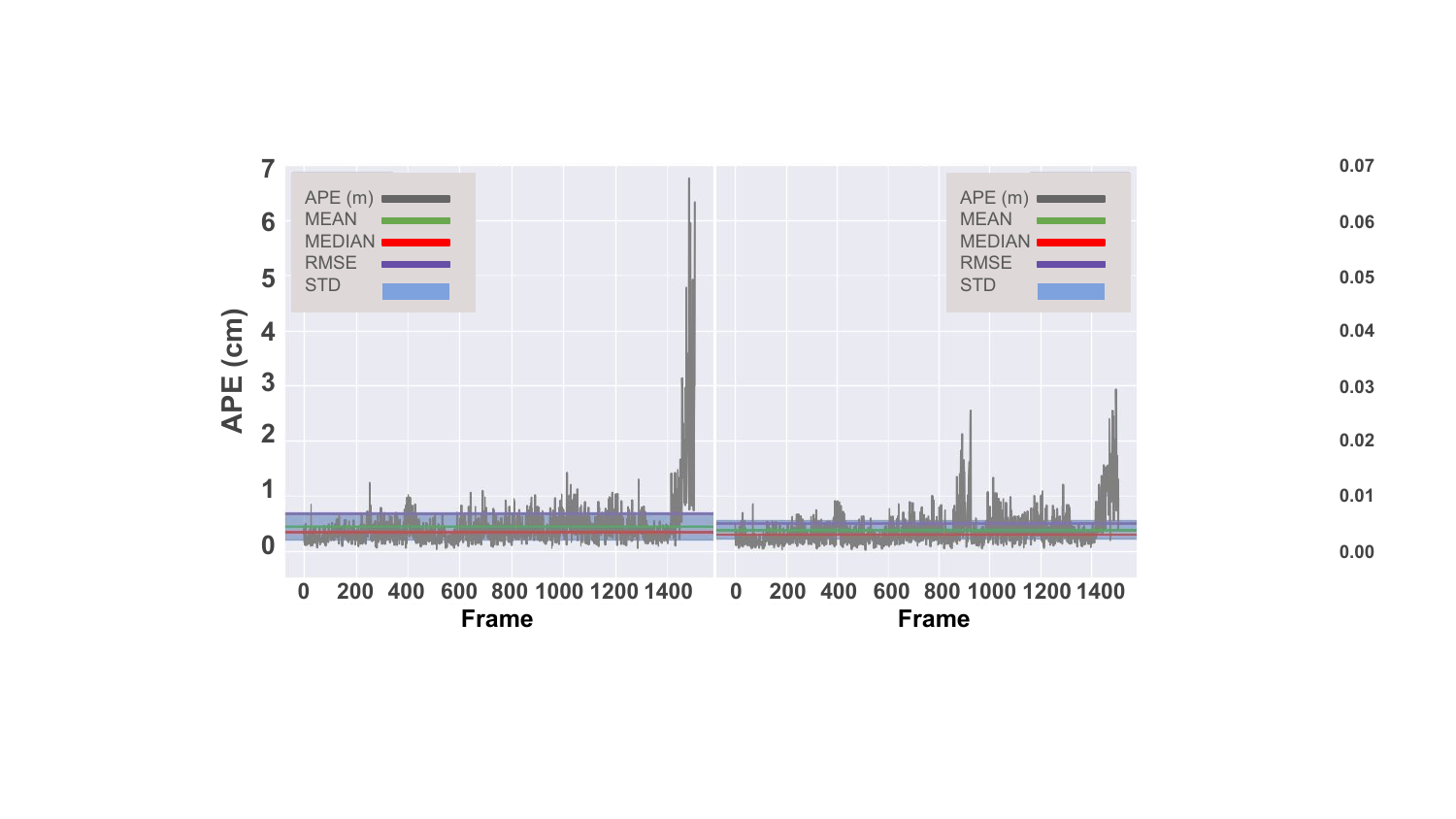}
    \caption{APE of translation in \textit{office0}. From left to right are \textcolor{black}{the results of Point-BA and PointLine-BA} methods.}
    \label{fig:office0-ape}
\end{figure}


\begin{table*}[]
    \centering
    \resizebox{\textwidth}{!}{
        \begin{tabular}{l|cccccccc}
            \toprule
            \multirow{3}{*}{Sequence} & \multicolumn{4}{c}{Initial Pose}   & \multicolumn{4}{c}{Co-visibility Graph Optimization}                                                                                                                                         \\
                                      & \multicolumn{2}{c}{Frame-to-Frame} & \multicolumn{2}{c}{Map-to-Frame}                     & \multicolumn{2}{c}{Point-BA} & \multicolumn{2}{c}{PointLine-BA}                                                                       \\
                                      & Translation (cm)                    & Rotation (deg)                                       & Translation (cm)              & Rotation (deg)                   & Translation (cm) & Rotation (deg) & Translation (cm) & Rotation (deg) \\ \hline
            livingroom1               & 14.2/12.3                        & 0.45/0.26                                          & 23.0/1.8                  & 0.49/0.21                      & 1.6/\textbf{1.0}     & 0.25/0.17    & \textbf{1.5}/\textbf{1.0}     & \textbf{0.23}/\textbf{0.15}    \\
            
            livingroom2               & 25.3/20.6                        & 0.73/0.20                                          & 8.1/3.9                  & 1.61/0.19                      & 1.4/0.6     & 0.53/0.13    & \textbf{0.7}/\textbf{0.5}     & \textbf{0.24}/\textbf{0.11}    \\
            
            office2                   & 20.8/19.2                        & 0.44/0.16                                          & 3.3/2.1                  & 0.66/0.14                      & 1.1/\textbf{0.9}     & 0.27/0.10    & \textbf{1.0}/\textbf{0.9}     & \textbf{0.18}/\textbf{0.09}    \\ 
            
            office3                   & 18.5/16.7                        & 0.58/0.18                                          & 5.3/2.3                  & 1.69/0.15                      & 1.3/\textbf{0.6}     & 0.52/0.11    & \textbf{0.9}/\textbf{0.6}     & \textbf{0.32}/\textbf{0.10}    \\
            
            hospital                  & 37.5/32.7                        & 0.41/0.23                                          & 19.1/14.9                  & 0.52/0.26                      & \textbf{7.7}/7.1     & \textbf{0.19}/0.11    & 8.3/\textbf{5.7}     & 0.28/\textbf{0.10}    \\ 
            
            carwelding                & 41.1/33.6                        & 0.41/0.24                                          & 7.5/5.9                  & 0.38/0.20                      & 3.5/3.1     & 0.15/0.10    & \textbf{2.5}/\textbf{2.3}     & \textbf{0.12}/\textbf{0.09}    \\
            stru\_texture\_near       & 32.7/29.3                        & 1.00/0.58                                          & 4.5/3.5                  & 1.79/0.56                      & 1.0/\textbf{0.8}     & 0.46/0.28    & \textbf{0.9}/\textbf{0.8}     & \textbf{0.38}/\textbf{0.23}    \\
            
            nostru\_texture\_near     & 88.3/72.7                        & 1.13/0.64                                          & 18.3/6.4                  & 5.76/0.88                      & 4.4/\textbf{3.0}     & 2.91/0.32    & \textbf{3.3}/3.1     & \textbf{0.58}/\textbf{0.25}    \\\hline
            corridor1                 & 216.4/148.9                        & 1.09/0.14                                          & 10.5/5.9                  & 1.09/0.14
                                      & 5.5/4.4                        & 0.31/0.10
                                      & \textbf{4.1}/\textbf{3.2}                        & \textbf{0.26}/\textbf{0.09}                                                                                                                                                                                  \\
                                      
            corridor2                 & 36.4/27.1                        & 0.34/0.10                                          & 52.1/24.9                  & 0.59/0.11                     & 13.9/13.9     & \textbf{0.16}/\textbf{0.07}
                                      & \textbf{6.1}/\textbf{6.0}                        & \textbf{0.16}/\textbf{0.07}                                                                                                                                                                                  \\
            box1                      & 7.6/5.5                        & 0.18/0.14                                          & 2.0/1.6                  & 0.18/0.13                      & \textbf{1.7}/\textbf{1.4}     & 0.16/0.12
                                      & 1.8/1.5                        & \textbf{0.15}/\textbf{0.11}                                                                                                                                                                                  \\
            box2                      & 73.9/57.6                        & 0.29/0.18                                          & 9.6/3.0                  & 0.40/0.18                      & 2.8/1.5     & 0.33/0.14
                                      & \textbf{2.0}/\textbf{1.4}                        & \textbf{0.24}/\textbf{0.13}                                                                                                                                                                                  \\
            \bottomrule
        \end{tabular}}
    \caption{\textcolor{black}{Comparison of translation (APE) RMSE/MEDIAN and rotation (RPE) RMSE/MEDIAN on the Open-Structure benchmark dataset. Results with the best accuracy are highlighted by \textbf{bold} font}. }
    \label{tab:quantitative_results}
\end{table*}

\begin{figure*}
    \centering
    \includegraphics[width=0.97\linewidth, trim = 0.8cm 2cm 0.55cm 0.8cm, clip]{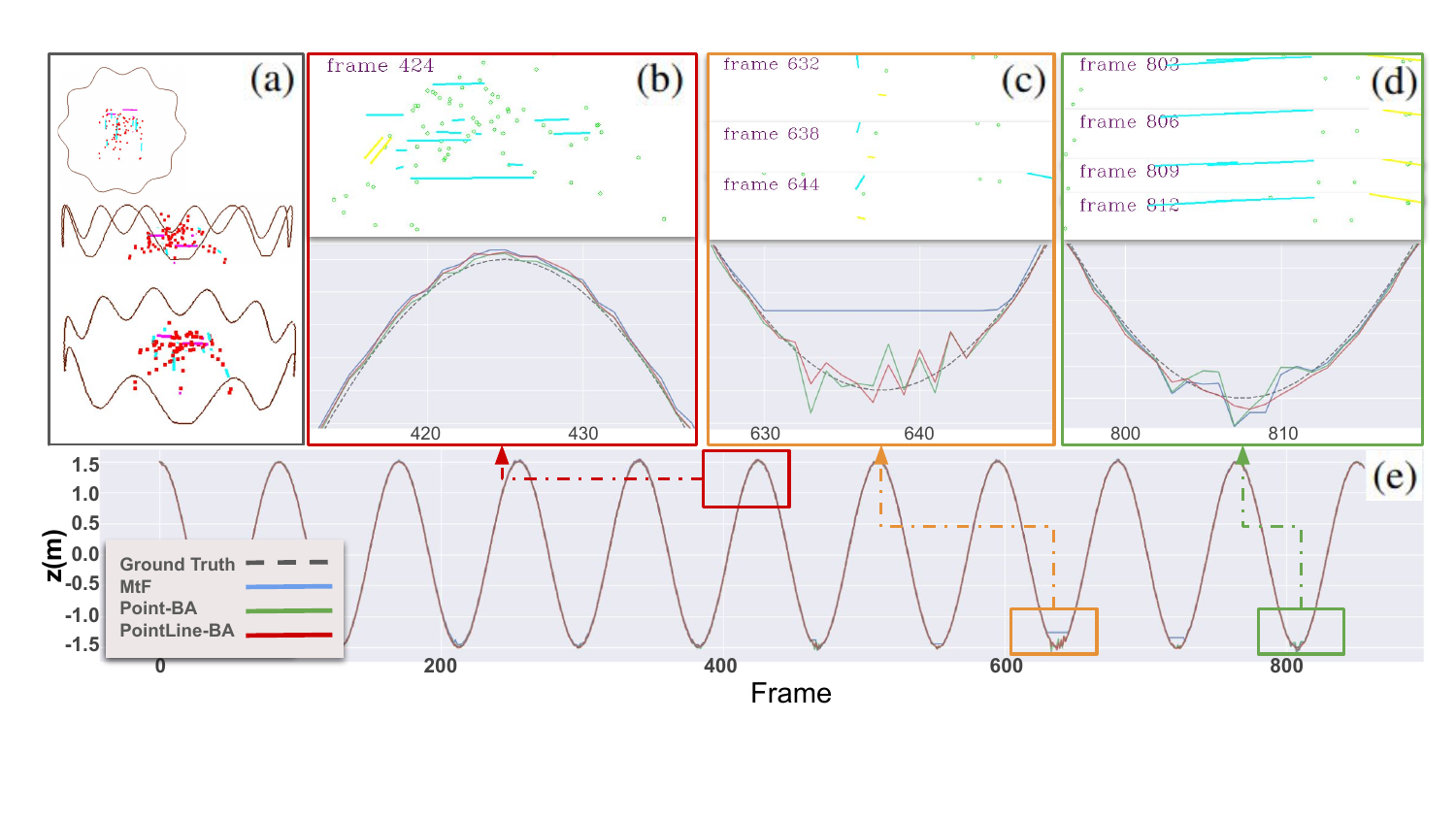}
    \caption{Comparison in trajectory results of \textit{MtF}, Point-BA, and PointLine-BA methods in the box2 sequence. (a) shows the simulated scene and robot trajectory from different viewpoints; (b)-(d) are zoom-in trajectories and corresponding measurements. Red rectangles show sufficient observation cases, while yellow and green rectangles highlight achieved trajectories in low-textured scenarios; (e) shows trajectories in the z-aix direction.}
    \label{fig:simu_rot}
\end{figure*}

\input{figure_rmse.tex}

\subsection{Real World Sequences}
\textcolor{black}{Table~\ref{tab:quantitative_results} presents the evaluation results for eight sequences of S-\uppercase\expandafter{\romannumeral1}, assessing two initial pose estimation approaches} (\textit{FtF} and \textit{MtF}) and two factor graph optimization methods (Point-BA and PointLine-BA). The \textit{MtF} tracking strategy demonstrates greater robustness than the \textit{FtF} method, despite both using the same core pose computation algorithm (EPnP~\cite{lepetit2009ep}). \textit{MtF} benefits from more 3D-2D pairs for pose estimation, particularly evident in sequences like \textit{carwelding} and \textit{office2}.

\textcolor{black}{The initial camera poses estimated by the \textit{MtF} method are then refined through a factor graph optimization module, leading to continuous improvements in both translation and rotation accuracy. For instance, in the \textit{office3} and \textit{stru\_texture\_near} sequences, Point-BA reduces errors to one-third of the initial \textit{MtF} estimates, highlighting the significant role of bundle adjustment modules in mitigating camera pose drift. In the factor graphs of Point-BA and PointLine-BA}, additional co-visible connections of line observations enhance the performance of point-only optimization in the \textit{livingroom2} sequence, reducing errors from $0.014$m to $0.007$m. \textcolor{black}{However, in sequences where sufficient points are detected, such as \textit{office2} and \textit{stru\_texture\_near}, additional line constraints do not significantly improve optimization results. Figure~\ref{fig:office0-ape} illustrates the \textcolor{black}{optimization results from Point-BA and PointLine-BA algorithms}, with PointLine-BA exhibiting greater robustness in RMSE (root mean square error) and STD (standard deviation) metrics}. Additional qualitative results for \textit{office0} are available in the supplementary video.

\subsection{Simulation Sequences}~\label{sec:private}
\textcolor{black}{In Figure \ref{fig:sphere_traj}, \ref{fig:box1_traj}, and \ref{fig:corridor2}, we observe different challenges posed by each sequence. \textit{sphere1} presents difficulties in orientation estimation due to minimal frame overlaps. On the other hand, \textit{corridor2} simulates robot motions in a corridor environment, where the movements are primarily pure translation with sharp rotation changes at corners. \textit{box1}, in contrast, offers a balanced mix of translation and rotation challenges, following a wave-shaped trajectory}.

Table \ref{tab:quantitative_results} presents the camera pose results for four simulation sequences. The \textit{MtF} strategy generally outperforms the relative-frame-based pose estimation method. However, in \textit{corridor2}, \textcolor{black}{the initial sparse map quality is poor, particularly in corner regions, leading to significant drift in the \textit{MtF} method towards the end of the trajectory, as depicted in Figure \ref{fig:corridor2-traj}. The convergence steps of Point-BA and PointLine-BA are illustrated in Figure \ref{fig:convergence}}.

\textcolor{black}{Figure \ref{fig:simu_rot} investigates the relationship between feature distribution and the robustness of camera pose estimation. When sufficient correspondences are detected, as shown in Figure \ref{fig:simu_rot}(b)}, the EPnP method~\cite{lepetit2009ep} based on 3D-2D observations \textcolor{black}{produces} acceptable results. \textcolor{black}{Figure \ref{fig:simu_rot}(c) demonstrates that refining the initial estimates based on factor graph optimization methods leads to improved results. Additionally, longer 2D lines enhance the robustness of trajectory optimization, as seen from the $803^{rd}$ frame to the $812^{th}$ frame}.


\textcolor{black}{Furthermore, we compare the convergence curves of Point-BA and PointLine-BA in the optimization process, as shown in Figure \ref{fig:convergence}. This comparison illustrates that the PointLine-BA method achieves lower residuals in the refined factor graph compared to Point-BA in the sequence \textit{corridor2}. Additionally, Figure \ref{fig:corridor2-traj} provides visualizations of the initial poses, optimized trajectories, and ground truth for the sequence}.

\section{Conclusion and Future Works}
This paper introduces Open-Structure, a new dataset benchmark to facilitate the fair and efficient evaluation of SLAM modules, where measurements, 3D landmarks, initial camera poses, structural relationships, and co-visibility factor graphs can be directly obtained.
\textcolor{black}{There are two types of sequences in our benchmark. The first category of scenes maintains the same observation and occlusion relationships as presented in image-based datasets. In the other type, simulation sequences are designed to include challenging motions and environments, providing an evaluation platform for evaluating the performance of methods in extreme cases.}

For future works, new parameterization and optimization strategies are expected to be explored, especially based on the proposed structural regularities, to achieve more accurate and robust pose estimation and reconstruction performance.


\bibliographystyle{IEEEtran}
\bibliography{root}

%
%
%
%
%
%

\end{document}